\pdfoutput=1

\documentclass[11pt]{article}

\usepackage{amsmath}

\usepackage[preprint]{acl}
\usepackage{booktabs}
\usepackage{hyperref}
\usepackage[table]{xcolor}
\usepackage{times}
\usepackage{latexsym}
\usepackage{multirow}
\usepackage{tabularx}
\usepackage{pifont}
\usepackage{tcolorbox}
\usepackage{enumitem}

\usepackage{multirow}      
\usepackage{booktabs}      
\usepackage{adjustbox}     
\usepackage{caption}       
\usepackage{geometry}      
\usepackage{amssymb}
\usepackage{graphicx}
\usepackage{subcaption}
\usepackage[T1]{fontenc}

\usepackage[utf8]{inputenc}

\usepackage[linesnumbered,ruled,vlined]{algorithm2e}
\usepackage{textcomp} 

\usepackage{microtype}

\usepackage{inconsolata}
\usepackage{xcolor}
\usepackage{graphicx}
\usepackage{framed}
\usepackage{pifont}
\definecolor{shadecolor}{rgb}{0.92,0.92,0.92}

\usepackage{graphicx}   
\usepackage{booktabs}   
\usepackage{wasysym}    

\PassOptionsToPackage{table}{xcolor} 
\usepackage{xcolor}

\definecolor{myblue}{HTML}{418BBF}
\definecolor{myred}{HTML}{DC4748}
\definecolor{mypink}{HTML}{E78BCB}
\definecolor{mycyan}{HTML}{3BCCDB}

\newcommand{\Approach}[1]{{HyperEdit}}

\title{HyperEdit: Unlocking Instruction-based Text Editing in LLMs via Hypernetworks}

\author{%
\textbf{Yiming Zeng}$^{1}$, 
\textbf{Jinghan Cao}$^{2}$, 
\textbf{Zexin Li}$^{3}$, 
\textbf{Wanhao Yu}$^{4}$,
\textbf{Zhankai Ye}$^{5}$, \\
\textbf{Dawei Xiang}$^{1}$, 
\textbf{Ting Hua}$^{6}$, 
\textbf{Xin Liu}$^{5}$, 
\textbf{Shangqian Gao}$^{5\dagger}$, 
\textbf{Tingting Yu}$^{1\dagger}$ \\[2mm]
$^{1}$University of Connecticut,\;
$^{2}$San Francisco State University,\\
$^{3}$University of California, Riverside,\;
$^{4}$University of North Carolina at Charlotte,\\
$^{5}$Florida State University,\;
$^{6}$University of Notre Dame\;
\\
}
\usepackage{float}
\usepackage[table,xcdraw]{xcolor} 
\definecolor{TaskBG}{RGB}{245,245,245}        
\definecolor{PrimaryBlue}{RGB}{0,102,204}     
\definecolor{PrimaryRed}{RGB}{204,0,0}        

\begin{document}
\maketitle
\begingroup
  \renewcommand\thefootnote{$\dagger$}
  \footnotetext{Corresponding authors}
\endgroup
\begin{abstract}
Instruction-based text editing is increasingly critical for real-world applications such as code editors (e.g., Cursor), but Large Language Models (LLMs) continue to struggle with this task. Unlike free-form generation, editing requires faithfully implementing user instructions while preserving unchanged content, as even minor unintended modifications can break functionality. Existing approaches treat editing as generic text generation, leading to two key failures: they struggle to faithfully align edits with diverse user intents, and they often over-edit unchanged regions.
We propose HyperEdit to address both issues. First, we introduce hypernetwork-based dynamic adaptation that generates request-specific parameters, enabling the model to tailor its editing strategy to each instruction. Second, we develop difference-aware regularization that focuses supervision on modified spans, preventing over-editing while ensuring precise, minimal changes. HyperEdit achieves a \textbf{9\%--30\%} relative improvement in BLEU on modified regions over state-of-the-art baselines, despite utilizing only 3B parameters. 



\end{abstract}

\section{Introduction}
Large Language Models (LLMs) have fundamentally transformed natural language processing (NLP) and its real-world applications~\cite{qu2025silmm, reguloGPT, wu2024auto, gao2025tomoe, feedback-bug, Genome-wide, chen-etal, concurrent}, significantly accelerating advancements in text generation, reasoning, and summarization~\cite{ravaut-etal-2024-context, song-etal-2024-finesure, li-etal-2024-improving-attributed, li2025learning, liu-etal-2025-llms-capable-custom, distillReason}. Recently, the editing task has been brought to people's attention~\cite{shu2024rewritelm, BeyondtheChat, raheja-etal-2023-coedit, zeng2025bridgingeditinggapllms}. In NLP, the editing task refers to refining an existing draft by applying minimal, targeted modifications rather than fully regenerating the text~\cite{iso2020,stahlberg-kumar-2020-seq2edits}. LLM-based editing has seen wide adoption in real-world applications such as Cursor for code editing, Grammarly for grammar correction, and Notion AI for collaborative document revision~\cite{NotionAI, Grammarly, CursorEditor}. Unlike traditional generation tasks, editing requires faithfully implementing user instructions while preserving unchanged content. In code editing tools like Cursor, even minor unintended modifications can break functionality~\cite{cassano2024editevaluatingabilitylarge}. Similarly, in grammatical error correction, the model must amend only erroneous tokens while leaving correct elements untouched~\cite{alhafni-habash-2025-enhancing}.
These examples underscore the fundamental constraints of editing: \textit{(1) Faithful adherence to user intent; (2) Preservation of unchanged context.}

However, as illustrated in Table~\ref{tab:baselines}, existing approaches fail to satisfy both constraints simultaneously. InstructGPT prioritizes instruction following but allows over-editing of unchanged regions~\cite{InstructGPT}. FineEdit benchmarks diverse editing tasks and incorporates edit differences, yet lacks explicit constraints during training to enforce minimal intervention~\cite{zeng2025bridgingeditinggapllms}. Similarly, code editing methods emphasize functional correctness while often overlooking local fidelity~\cite{cassano2024editevaluatingabilitylarge}.
These limitations raise a key question: \textit{can we design a method that simultaneously achieves intent alignment and local fidelity for reliable editing performance?}

\begin{table}[!tbp]
  \centering
  \renewcommand{\arraystretch}{1.3} 
  \resizebox{0.48\textwidth}{!}{
    \begin{tabular}{ccccccc}
    \toprule
    \textbf{Baseline}  & \textbf{Constraint 1} & \textbf{Constraint 2} \\
    \midrule
{FineEdit~\cite{zeng2025bridgingeditinggapllms}}  &  \LEFTcircle &  \LEFTcircle \\
{CODA~ \cite{cassano2024editevaluatingabilitylarge}}  &   \CIRCLE & \Circle   \\
{EDITCODER~ \cite{alhafni-habash-2025-enhancing}} & \LEFTcircle & \CIRCLE    \\
InstructGPT~\cite{InstructGPT} & \CIRCLE  & \Circle \\
\midrule
\Approach{} & \CIRCLE  & \CIRCLE   \\
\Approach{}$_{Pro}$ & \CIRCLE  & \CIRCLE \\
    \bottomrule
    \end{tabular}%
    
    }
  \caption{Existing baselines and their supported constraints on the editing task. $\CIRCLE$ represents support, $\LEFTcircle$ represents partial support, and $\Circle$ represents no support. }
  \vspace{-5mm}
  \label{tab:baselines}%
\end{table}%

To answer this question, we propose HyperEdit, a framework that integrates two key designs to address both constraints: (i) a hypernetwork-based dynamic modeling mechanism that generates request-specific parameters to adapt editing strategies to diverse user intents, and (ii) drawing inspiration from localization fidelity in computer vision~\cite{zhao2024balf}, we design a difference-aware regularization that focuses supervision on modified spans, ensuring precise modifications with minimal intervention. By enabling adaptive intent alignment and targeted content preservation, HyperEdit achieves state-of-the-art editing performance. 
Our contributions can be summarized as follows: 
\begin{itemize}[leftmargin=1em, itemsep=0pt, parsep=0pt, topsep=0pt]  
    \item We propose a hypernetwork-based dynamic adaptation mechanism that generates request-specific low-rank parameters, enabling the model to tailor its editing behavior to diverse instructions without retraining. To our knowledge, this is the first work to introduce dynamic parameter generation for instruction-based text editing.

    \item We develop a difference-aware regularization loss that applies targeted supervision to edited spans, guiding the model to focus on necessary modifications while keeping unchanged content.

    \item Extensive experiments demonstrate that HyperEdit achieves substantial improvements over existing methods: approximately 9\%--30\% relative improvement over the state-of-the-art editing model FineEdit-Pro, and 50\% relative improvement over the 14B-parameter Qwen-2.5-14B-Instruct, while using only 3B parameters. 
\end{itemize}

\begin{figure*}[!htbp]
    \centering
    \includegraphics[width=0.9\textwidth]{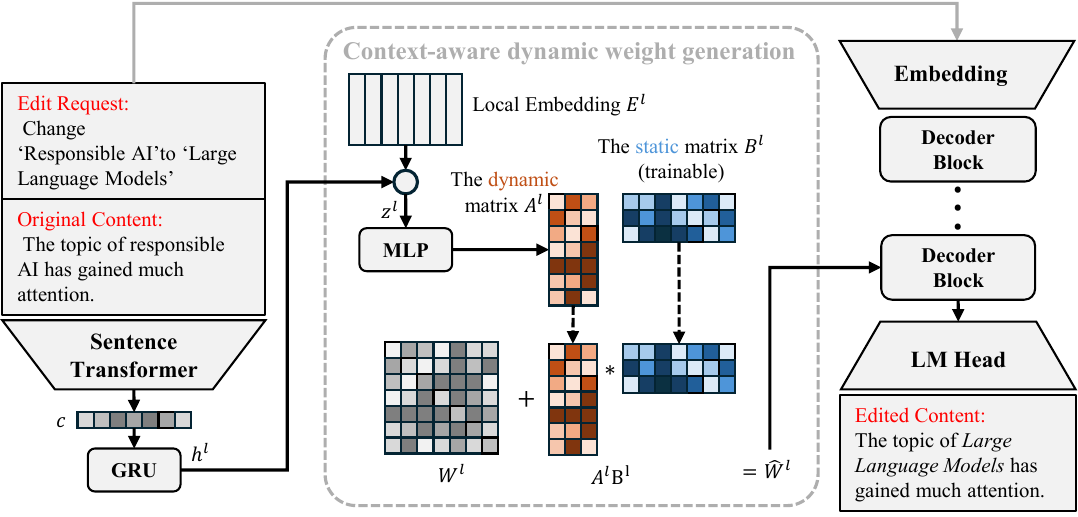}
    \caption{The process of context-aware dynamic weight generation. The hypernetwork will generate the dynamic weight matrix $\mathbf{A}^l$ given the edit request and the original text. The original model weights $\mathbf{W}^l$ are then modified to produce $\hat{\mathbf{W}}^l=\mathbf{W}^l+\mathbf{A}^l\mathbf{B}^l$ to capture context-specific requirements from the prompt.
    }

    \label{fig:example-pdf}
\end{figure*}

\section{Related Work}

\subsection{LLM Editing Task}
Instruction tuning has enabled LLMs to perform editing in more diverse and user-driven scenarios. Models such as InstructGPT~\cite{InstructGPT}, RewriteLM~\cite{shu2024rewritelm}, and DocMEdit~\cite{zeng-etal-2025-docmedit} exhibit strong abilities in following user commands but sometimes suffer from over-editing or altering irrelevant content.

Although instruction-tuned LLMs show promise in editing, existing approaches are often fragmented across tasks and domains. Recent studies therefore propose unified methods to address this limitation. FineEdit introduces InstrEditBench, a structured benchmark spanning multiple domains to assess instruction-based edits~\cite{zeng2025bridgingeditinggapllms}. CoEdIT develops an instruction-tuned editing system that generalizes across diverse tasks~\cite{raheja-etal-2023-coedit}, and its multilingual extension mEdIT extends this paradigm to multiple languages~\cite{raheja-etal-2024-medit}. XATU focuses on fine-grained and explainable editing benchmarks by explicitly annotating editing rationales~\cite{zhang-etal-2024-xatu}.

On the dataset side, Beemo captures real-world human post-edits over machine-generated text, enabling analysis of how expert edits differ from purely automated ones~\cite{artemova-etal-2025-beemo-custom}. Research has also examined LLMs for code editing: Cassano et al.~\cite{cassano2024editevaluatingabilitylarge} assessed LLMs on code editing instructions, and Fan et al.~\cite{fan2025exploring} studied realistic scenarios like commit message generation and code reviews.

Existing studies mainly focus on building benchmarks or designing systems for broad task coverage, but they often lack mechanisms to ensure both intent alignment and local fidelity. Our method belongs to the instruction-based editing category and introduces hypernetwork-driven dynamic adapters together with a difference-aware loss to achieve precise and minimally invasive edits.

\subsection{Hypernetwork Applications}
Hypernetworks are a class of neural network methods that improve generalization and adaptability by dynamically generating target model parameters. Early work has demonstrated the advantages of hypernetworks in few-shot learning and continual learning~\cite{ha2017hypernetworks, vonoswald2020continual}. By sharing a transferable parameter generator, they reduce optimization time and alleviate overfitting. Recent research has further expanded the application of hypernetworks in implicit neural representations (INRs), including 3D object modeling~\cite{sen2024hypnerf}, video and image stylization~\cite{maiya2024latentinr}, and video decomposition~\cite{pilligua2025hypernvd}.

While hypernetworks have proven effective in vision tasks, they remain underexplored in NLP. Editing is inherently instruction-sensitive: the model must adapt its response to each request while preserving unrelated content. Static adapters and traditional fine-tuning fail to capture this variability, as they rely on the same parameters across all inputs. Hypernetworks overcome this limitation by generating context-dependent weights, enabling fine-grained and flexible adaptation crucial for instruction-based editing.

\section{Method}

\subsection{Problem Formulation}

We define the instruction-based editing task as transforming an original text \(T_{\text{orig}}\) into an edited text \(T_{\text{edit}}\) under the guidance of an editing instruction \(I_{\text{edit}}\). The objective is to ensure that \(T_{\text{edit}}\) faithfully incorporates the modifications specified by \(I_{\text{edit}}\), while preserving the content and structure of \(T_{\text{orig}}\) wherever changes are not required.
Formally, the task can be expressed as a conditional mapping:
\begin{equation}
    T_{\text{edit}} = f_{\text{LLM}}\bigl(T_{\text{orig}}, I_{\text{edit}}; \theta\bigr),
\end{equation}
where \(\theta\) denotes the learnable parameters, and \(f_{\text{LLM}}\) is instantiated by a language model that generates the edited text given the input pair \((T_{\text{orig}}, I_{\text{edit}})\).

Instruction-based editing requires the model to perform minimal yet precise modifications: unchanged portions of \(T_{\text{orig}}\) should be preserved as much as possible, while only the regions targeted by \(I_{\text{edit}}\) are modified. This makes the task more constrained than general text generation, as the model must balance two competing objectives: (i) faithfully implementing the instruction \(I_{\text{edit}}\), and (ii) avoiding unnecessary alterations to \(T_{\text{orig}}\).

In practice, editing requests vary widely in type and complexity, requiring the model to dynamically adapt its behavior to different instructions. To this end, we propose a \textit{hypernetwork-based dynamic adaptation} scheme that generates request-specific parameters on the fly, enabling tailored editing strategies for each input. However, standard likelihood objectives may still under-emphasize the regions where edits actually occur, since unchanged tokens dominate the sequence. To address this, we further introduce a \textit{difference-aware regularization} that places targeted supervision on modified spans, ensuring faithful and minimal edits.

\subsection{Training Process}
Our training process incorporates two key mechanisms: a hypernetwork that generates request-specific low-rank parameters for dynamic adaptation, and a difference-aware regularization loss that focuses supervision on modified spans.

\subsubsection{HyperNetwork for Dynamic Low-Rank Adaptation \label{sec:3.2.1}}

To better capture diverse editing intents in a flexible manner, we adopt a hypernetwork-based dynamic parameter generation scheme illustrated in Figure~\ref{fig:example-pdf}. Specifically, following the LoRA paradigm~\cite{hu2022lora}, the hypernetwork produces input-dependent, low-rank adaptation matrices for selected layers of the frozen LLM. This design enables the model to disentangle general editing patterns from instruction-specific contexts, and to construct specialized adapters tailored to each editing request.

The process begins by encoding the edit instruction $I_{\text{edit}}$ and the original text
chunk $T_{\text{orig}}$ into a single context embedding $\mathbf{c} \in \mathbb{R}^{d_{\text{emb}}}$ using a pre-trained, frozen sentence encoder~\cite{reimers2019sentence}, which provides a rich semantic representation of the editing task. Formally, we concatenate the two inputs and feed them into the encoder:
\begin{equation}
x = [I_{\text{edit}} ; T_{\text{orig}}], 
\quad \mathbf{c} = \mathrm{ST}(x) \in \mathbb{R}^{d_{\text{emb}}},
\end{equation}
where $\mathrm{ST}(\cdot)$ denotes the pre-trained sentence transformer encoder that maps the concatenated input $x$ into the embedding space.

The hypernetwork consists of three components: a \textbf{GRU} that encodes the context vector $\mathbf{c}$ obtained from the sentence transformer for each weight matrix, a \textbf{local embedding} that stores $d_{\text{in}}^l$ vectors of rank $r$, and an \textbf{MLP} that generates the dynamic low-rank weights $\mathbf{A}^l$ based on the local embeddings and the GRU outputs.

More specifically, the formulation of GRU is as follows:
\begin{equation}
        \mathbf{h}^{l} = \mathrm{GRU}\!\left(\mathbf{c}, \mathbf{h}^{l-1}\right),
\end{equation}
where $\mathbf{h}^{l}$ is the encoded $\mathbf{c}$ for the $l$-th weight matrix. The local embedding $E^{l} \in \mathbb{R}^{d_{\text{in}}^l \times r}$ stores learnable codes for each dimension in $d_{\text{in}}^l$. $E^{l}$ is a static local embedding, and as a result, we need to fuse the local embedding with the request-based encoding $\mathbf{h}^{l}$ from $\textbf{c}$. To achieve this, we concatenate $E^{l}$ and $\mathbf{h}^{l}$ together and apply layer normalization (LN)~\cite{ba2016layer} and GeLU~\cite{hendrycks2016gaussian} before feeding them into the final MLP layer:
\begin{equation}
     \mathbf{A}^l = \mathrm{MLP}( \mathrm{GeLU}(\mathrm{LN}(\mathbf{z}^{l})))
\end{equation}
where $\mathbf{z}^l = [E^{l}; \mathrm{expand}(\mathbf{h}^{l})]$, and $\mathbf{h}^{l}\in \mathbb{R}^{d_{\text{h}}}$ is first expanded to $d_{\text{in}}^l\times d_{\text{h}}$, $\mathbf{z}^l\in \mathbb{R}^{d_{\text{in}}^l\times(d_{\text{h}}+r)}$, and $W_{\mathrm{MLP}}^l\in \mathbb{R}^{(r+d_{\text{h}})\times r}$. The final size of $\mathbf{A}^l$ is the same as the local embedding $E^l$: ${d_{\text{in}}^l \times r}$. The function of $\mathrm{MLP}(\cdot)$ is to fuse information from the static local embeddings and the adaptive input $\mathbf{h}^l$. Consequently, the generated $\mathbf{A}^l$ becomes conditionally dependent on the editing request and the corresponding text. During this process, we generate only $\mathbf{A}^l$ while keeping $\mathbf{B}^l$ input-independent, since $\mathbf{B}^l$ is initialized as an all-zero matrix in LoRA.

The additional learnable parameters introduced by our design are quite small. The parameters of the GRU can be neglected due to its small hidden size. Compared to the original LoRA, the extra trainable parameters are $r \times (d_{\text{h}} + r)$, where $d_{\text{h}}$ is typically set to a small value such as 32 or 64.

At train time, given an input $\mathbf{X}^l \in \mathbb{R}^{T\times d_{\text{in}}^l}$, the adapted output of the $l$-th weight matrix is computed as:
\begin{equation}
    \mathcal{F}^l =  \mathbf{X}^lW^{l}
    + \Delta \mathcal{F}^l , 
    \quad 
    \Delta \mathcal{F}^l  = \frac{\alpha}{r}\, 
    \mathbf{X}^l\mathbf{A}^{l}\mathbf{B}^{l},
\end{equation}
where $\alpha$ is a constant scaling factor.
This formulation keeps the original weight matrix $W^{l}$ frozen 
while dynamically constructing a context-dependent $\Delta \mathcal{F}^l$ for each editing request.

During training, both the sentence transformer and the LLM are kept frozen. The learnable parameters of our method are denoted as $\Theta = [\theta_{\mathrm{GRU}}, \theta_{\mathrm{MLP}}, E, \mathbf{B}]$, where $\theta_{\mathrm{GRU}}$ represents the parameters of the GRU, $\theta_{\mathrm{MLP}}$ corresponds to the parameters of all MLP layers and LNs, $E$ includes all local embeddings $E^l$, and $\mathbf{B}$ includes the parameters of all $\mathbf{B}^l$ matrices.

\subsubsection{Difference-Aware Regularization} 

In addition to standard supervised fine-tuning, we introduce a \emph{difference-aware loss} that emphasizes supervision on tokens deviating from the original text. Given the target sequence \(T_{\mathrm{edit}}\) and the reference context \(T_{\mathrm{orig}}\), we compute token-level alignment via the longest common subsequence (LCS), which identifies positions where tokens differ between the two sequences. This yields a binary mask \(m_{1:T}\) where \(m_t = 1\) indicates that position \(t\) contains an insertion or replacement, and \(m_t = 0\) indicates unchanged tokens. The loss is then applied exclusively to the modified positions. Formally, for the token sequence \(y_{1:T}\) of \(T_{\mathrm{edit}}\) and mask \(m_{1:T}\), the loss is defined as:
\begin{equation}
\mathcal{L}_{\mathrm{diff}} = - \frac{1}{\sum_{t} m_t} \sum_{t=1}^{T} m_t \,\log p\bigl(y_t \mid y_{<t}, T_{\mathrm{orig}}, I_{\mathrm{edit}}\bigr).
\end{equation}

This regularization loss guides the model to focus learning on modified regions while reducing penalties on unchanged text. The overall training objective combines \(\mathcal{L}_{\mathrm{diff}}\) with the supervised fine-tuning loss~\cite{InstructGPT}:
\begin{equation} \label{eq:overall_obj}
\min_{\Theta}\ \mathcal{L}_{\mathrm{sft}} + \lambda\,\mathcal{L}_{\mathrm{diff}},
\end{equation}
where \(\Theta\) is introduced in Section~\ref{sec:3.2.1}, \(\mathcal{L}_{\mathrm{sft}}\) is the language modeling loss applied to the edited response, and \(\lambda\) balances the two objectives. Empirically, we find \(\lambda = 1\) performs robustly across diverse editing domains.

\subsection{Inference Process}
At inference time, the hypernetwork generates request-specific adapters for each editing task.
Given a new editing request with instruction $I_{\text{edit}}$ and original text $T_{\text{orig}}$,
we first encode them into a context vector $\mathbf{c}$ using the frozen sentence transformer (as in training).
The hypernetwork then processes $\mathbf{c}$ to generate dynamic adapter matrices $\mathbf{A}^l$ for each target layer $l$.
These are combined with the static matrices $\mathbf{B}^l$ to form the low-rank updates:
\begin{equation}
\Delta W^l = \tfrac{\alpha}{r}\, \mathbf{A}^l \mathbf{B}^l.
\end{equation}

During the LLM forward pass, each layer $l$ applies the adapted weights.
For an input $\mathbf{X}^l$ and frozen base weight $W^l$, the output is computed as:
\begin{equation}
\mathcal{F}^l = \mathbf{X}^l(W^l + \Delta W^l).
\end{equation}
Note that, the base LLM parameters $W^l$ remain frozen throughout; only the dynamically generated $\Delta W^l$ varies across different editing requests, enabling request-specific adaptation without modifying the base model.
Since the additional computation involves only the lightweight hypernetwork forward pass and
low-rank matrix multiplications, the method remains efficient and scalable
across diverse editing scenarios.




\subsection{Evaluation Metrics}

We evaluate model performance with \textit{Diff-BLEU} and \textit{Diff-ROUGE-L},
two span-focused variants of standard text similarity metrics.

While BLEU~\cite{papineni2002bleu} and ROUGE-L~\cite{lin2004rouge} compute similarity over the entire output, our metrics restrict computation
to the \emph{edited spans} identified by LCS-based alignment between the original
and target text. This design better captures editing quality: since unchanged tokens typically dominate the output, standard metrics can yield high scores even when the model fails to make correct edits. By focusing only on modified regions, our metrics directly measure whether the model implements the intended modifications correctly.

\paragraph{Diff-BLEU.}
To evaluate the quality of the specific modifications, we calculate BLEU scores exclusively on the edited spans.
Let $Y_{\text{edit}}^{\text{span}} = (y_1, \ldots, y_M)$ denote the sequence of reference tokens extracted from the target side of the edits (i.e., capturing the content of \textit{insertions} and the \textit{new} components of \textit{replacements}).
Correspondingly, let $\hat{Y}_{\text{edit}}^{\text{span}} = (\hat{y}_1, \ldots, \hat{y}_N)$ denote the sequence of tokens generated by the model within the predicted edit spans.
\textit{Diff-BLEU} is defined as:
\begin{equation}
\mathit{Diff\text{-}BLEU} = BLEU\big(Y_{\text{edit}}^{\text{span}}, \hat{Y}_{\text{edit}}^{\text{span}}\big).
\end{equation}
This metric assesses whether the model accurately generates the intended modifications, focusing solely on the changed content rather than the unchanged context.

This measures the n-gram precision of modifications, evaluating whether the edited
tokens are accurate and well-formed.

\paragraph{Diff-ROUGE-L.}
Similarly, \textit{Diff-ROUGE-L} computes the F1-score based on the Longest Common Subsequence (LCS). 
Let $L$ denote the length of the LCS between the reference edited spans $Y_{\text{edit}}^{\text{span}}$ and the model's predicted spans $\hat{Y}_{\text{edit}}^{\text{span}}$.
We define recall $R = L / |Y_{\text{edit}}^{\text{span}}|$ and precision $P = L / |\hat{Y}_{\text{edit}}^{\text{span}}|$, where $|\cdot|$ represents the number of tokens.
The metric is calculated as:
\begin{equation}
\mathit{Diff\text{-}ROUGE\text{-}L} = \frac{2 \cdot P \cdot R}{P + R}
\end{equation}

This measures the coverage of edits, capturing whether the model produces the essential changes even under minor lexical variations. We discuss the details of edit-span extraction in Appendix ~\ref{app:span_extraction}.
\section{Evaluation}
\subsection{Experimental Setup}
\paragraph{Datasets}
We conduct experiments on InstrEditBench~\cite{zeng2025bridgingeditinggapllms}, which integrates instruction-based editing tasks from four domains: LaTeX, programming code, Wikipedia text, and domain-specific languages (DSL). Following FineEdit, we use a 90/10 train-test split. Detailed hyperparameters are provided in the appendix.

\paragraph{Models and Baselines}
We implement HyperEdit on two base models: LLaMA-3.2-3B~\cite{dubey2024llama} and Qwen-2.5-3B-Instruct~\cite{qwen2025qwen25technicalreport}, both fine-tuned on InstrEditBench with our proposed hypernetwork and difference-aware regularization.

We compare against two categories of baselines: (1) \textit{Editing-specialized models}: FineEdit-Pro and FineEdit-X~\cite{zeng2025bridgingeditinggapllms}, which are specifically designed for instruction-based editing; (2) \textit{General-purpose LLMs}: LLaMA-3.2-1B, LLaMA-3.2-3B, LLaMA-3.1-8B-Instruct~\cite{dubey2024llama}, Qwen-2.5-3B-Instruct, Qwen-2.5-14B-Instruct~\cite{qwen2025qwen25technicalreport}, and Qwen-3-8B~\cite{yang2025qwen3}, evaluated in zero-shot or instruction-tuned settings.

\begin{table*}[!htbp]
  \centering
  \renewcommand{\arraystretch}{1.2}
  \resizebox{\textwidth}{!}{
    \begin{tabular}{l|ccccc|ccccc}
      \toprule
      \multirow{2}{*}{\textbf{Model/Type}} & 
      \multicolumn{5}{c|}{\textbf{Diff-BLEU}} & 
      \multicolumn{5}{c}{\textbf{Diff-ROUGE-L}} \\
      \cmidrule(lr){2-6} \cmidrule(lr){7-11}
      & \textbf{Code} & \textbf{LaTeX} & \textbf{DSL} & \textbf{Wiki} & \textbf{Overall} 
      & \textbf{Code} & \textbf{LaTeX} & \textbf{DSL} & \textbf{Wiki} & \textbf{Overall} \\
      \midrule
      Llama-3.2-1B & 0.0366 & 0.0886 & 0.0272 & 0.2573 & 0.1139 & 0.1631 & 0.2424 & 0.0800 & 0.3769 & 0.2156 \\
      Llama-3.2-3B & 0.0568 & 0.1455 &  0.0523 & 0.2746 & 0.1323 & 0.1677 & 0.2492 & 0.1045 & 0.3830 & 0.2261 \\
      Llama-3.1-8B-Instruct & 0.0335 & 0.1244 & 0.0214 & 0.2523 & 0.1079 & 0.1123 & 0.2194 & 0.0519 & 0.3504 & 0.1835 \\
      LLama-3.1-13B-Instruct & 0.0286 & 0.0729 & 0.0119 & 0.2458 & 0.0898 & 0.0983 & 0.1379 & 0.0332 & 0.3396 & 0.1523 \\
      \midrule
      Qwen-2.5-3B-Instruct  & 0.0280 & 0.0929 & 0.0187 & 0.2791 & 0.1047 & 0.0840 & 0.1695 & 0.0472 & 0.3897 & 0.1726 \\
      Qwen-3-8B & 0.0320 & 0.0827 & 0.0171 & 0.2923 & 0.1060 & 0.0991 & 0.1463 & 0.0417 & 0.3992 & 0.1716 \\
      Qwen-2.5-14B-Instruct & 0.0281 & 0.0928 & 0.0194 & 0.2722 & 0.1031 & 0.0847 & 0.1623 & 0.0471 & 0.3814 & 0.1689 \\
      \midrule
      FineEdit-X  & 0.0483 & 0.1176 & 0.0389 & 0.3499 & 0.1387 & 0.1600 & 0.2101 & 0.0867 & 0.4536 & 0.2276 \\
      FineEdit-Pro & 0.0497 & 0.1278 & 0.0356 & \textbf{0.3617}& 0.1437 & 0.1668 & 0.2295  & 0.0674 & 0.4744 & 0.2345 \\
      \midrule
      \rowcolor{yellow!20} HyperEdit (llama3.2-3B) & 0.0515 & 0.1544 & \textbf{0.0693} & 0.3043 & 0.1449 & 0.1569 & 0.2632 & 0.1114 & 0.4123 & 0.2360 \\
      \rowcolor{yellow!20} HyperEdit (qwen-2.5-3B) & 0.0492 & 0.1706 & 0.0535 & 0.3527 & \textbf{0.1565} & 0.1539 & 0.2803 & 0.1014 & \textbf{0.4792} & 0.2537 \\

       \rowcolor{yellow!20} HyperEdit-Pro (qwen-2.5-3B) & \textbf{0.1714} & \textbf{0.2432} & 0.0472 & 0.2860 & \textbf{0.1870} & \textbf{0.2880} & \textbf{0.3653} & \textbf{0.1137} & 0.4245 & \textbf{0.2979} \\
       
      \bottomrule
    \end{tabular}
  }
  \caption{Comparison of models on Diff-BLEU and Diff-ROUGE-L across domains. Models are grouped by family and ordered by size.}
  \label{tab:main}
\end{table*}

\subsection{Overall Effectiveness}

We evaluate HyperEdit under both single-turn and multi-turn settings. The single-turn version processes one instruction per instance, while the multi-turn version contains several editing requests applied sequentially. Each edit builds upon the model’s previous output, simulating a realistic iterative editing process.

\paragraph{Single-turn Experiments}
Table \ref{tab:main} reports the overall results on InstrEdit-Bench across Code, LaTeX, DSL, and Wikipedia. General-purpose models exhibit limited editing capability, with most scoring below 0.11 on Diff-BLEU. Among them, LLaMA-3.2-3B performs best (0.1323 Diff-BLEU, 0.2261 Diff-ROUGE-L), yet even larger instruction-tuned models like Qwen-2.5-14B-Instruct (0.1031 / 0.1689) struggle with fine-grained editing. This suggests that standard instruction-following training is insufficient for precise, minimal-intervention edits.

Editing-specialized models show clear improvements. The FineEdit family substantially outperforms general-purpose LLMs, with FineEdit-Pro achieving 0.1437 Diff-BLEU and 0.2345 Diff-ROUGE-L. This confirms that task-specific training enhances editing quality.

HyperEdit achieves the strongest performance through dynamic adaptation. Our best variant (Qwen-2.5-3B) obtains 0.1565 Diff-BLEU and 0.2537 Diff-ROUGE-L—approximately 9\% relative improvement over FineEdit-Pro. Notably, HyperEdit accomplishes this with only 3B parameters while outperforming models 4-5× larger. The LLaMA-based variant (0.1449 / 0.2360) similarly surpasses all baselines, demonstrating the approach's generality. These gains stem from the hypernetwork's ability to generate request-specific parameters that adapt editing strategies to diverse instructions while maintaining local fidelity.

Domain-specific results reveal where dynamic adaptation provides the greatest benefit. On structured domains requiring precise modifications—LaTeX (0.1706 / 0.2803) and DSL (0.0693 / 0.1114)—HyperEdit achieves the strongest gains, benefiting from its ability to tailor editing patterns to domain-specific syntax. On Wikipedia, a more free-form domain, HyperEdit still substantially outperforms FineEdit-Pro (0.3527 vs. 0.3617 on BLEU; 0.4792 vs. 0.4744 on ROUGE-L), indicating that dynamic modeling helps even when edits involve stylistic or semantic changes rather than structural ones.

In addition, we further introduce HyperEdit-Pro(Qwen-2.5-3B), which extends the sliding window to 4096 tokens and removes truncation constraints to preserve the entire input sequence. This configuration achieves the highest performance of 0.1714 Diff-BLEU and 0.2432 Diff-ROUGE-L. These findings indicate that input integrity is critical for instruction-based editing. Preserving the full context enables the model to accurately capture long-range dependencies and maintain structural coherence, which are often compromised by standard truncation strategies.

Overall, HyperEdit demonstrates that combining hypernetwork-based dynamic adaptation with difference-aware regularization effectively addresses both fundamental editing constraints: faithful intent alignment and preservation of unchanged content. The consistent improvements across diverse domains and editing scenarios validate the effectiveness of our approach.

\paragraph{Multi-turn Experiments}

Multi-turn editing presents additional challenges beyond single-turn scenarios: models must keep consistency across sequential edits while avoiding error accumulation as modifications build upon previous outputs. Table~\ref{tab:diff-multiturn} shows results in this challenging setting.

General LLMs struggle with iterative editing. LLaMA-3.2-3B obtains 0.1770 Diff-BLEU and 0.2982 Diff-ROUGE-L, while larger models like Qwen-3-8B (0.1711 / 0.2649) perform comparably or worse. These results indicate that standard instruction-following training provides insufficient mechanisms for tracking cumulative changes and preserving local fidelity across editing rounds.

\begin{table*}[!tbp]
  \centering
  
  \renewcommand{\arraystretch}{1.2}
  \resizebox{1.0\textwidth}{!}{
    \begin{tabular}{|l|cc|cc|cc|cc|cc|}
      \hline
      \textbf{Model} 
        & \multicolumn{2}{c|}{\textbf{Code}}
        & \multicolumn{2}{c|}{\textbf{LaTeX}}
        & \multicolumn{2}{c|}{\textbf{DSL}}
        & \multicolumn{2}{c|}{\textbf{Wiki}}
        & \multicolumn{2}{c|}{\textbf{Overall}} \\
      \cline{2-11}
      & \textbf{BLEU} & \textbf{ROUGE}
      & \textbf{BLEU} & \textbf{ROUGE}
      & \textbf{BLEU} & \textbf{ROUGE}
      & \textbf{BLEU} & \textbf{ROUGE}
      & \textbf{BLEU} & \textbf{ROUGE} \\
      \hline
      HyperEdit [without diff and hypernet]
        & 0.0497 & 0.1668
        & 0.1278 & 0.2295
        & 0.0356 & 0.0674
        & 0.3617 & 0.4744
        & 0.1437 & 0.2345 \\
      \hline
      HyperEdit [without diff]
        & \textbf{0.0504} & 0.1628
        & \textbf{0.1339} & \textbf{0.2346}
        & \textbf{0.0521} & \textbf{0.0954}
        & 0.3423 & 0.4551
        & \textbf{0.1447} & \textbf{0.2370} \\
      \hline
      HyperEdit 
        & 0.0492 & 0.1539
        & \textbf{0.1706} & \textbf{0.2803}
        & \textbf{0.0535} & \textbf{0.1014}
        & 0.3527 & \textbf{0.4792}
        & \textbf{0.1565} & \textbf{0.2537} \\
      \hline
    \end{tabular}
  }
  \caption{Ablation study results on Diff-BLEU and Diff-ROUGE across Code, LaTeX, DSL, and Wiki datasets. BLEU stands for Diff-BLEU, while ROUGE stands for Diff-ROUGE. Values in \textbf{bold} exceed the corresponding base scores.}
  \label{tab:ablation}
\end{table*}
\begin{table*}[!htbp]
  \centering
  \renewcommand{\arraystretch}{1.2}
  \resizebox{\textwidth}{!}{
    \begin{tabular}{l|ccccc|ccccc}
      \toprule
      \multirow{2}{*}{\textbf{Model/Type}} & 
      \multicolumn{5}{c|}{\textbf{Diff-BLEU}} & 
      \multicolumn{5}{c}{\textbf{Diff-ROUGE-L}} \\
      \cmidrule(lr){2-6} \cmidrule(lr){7-11}
      & \textbf{Code} & \textbf{LaTeX} & \textbf{DSL} & \textbf{Wiki} & \textbf{Overall} 
      & \textbf{Code} & \textbf{LaTeX} & \textbf{DSL} & \textbf{Wiki} & \textbf{Overall} \\
      \midrule
      Llama-3.2-1B & 0.08 & 0.1991 & 0.0836 & 0.3377 & 0.1735 & 0.2093 & 0.3252 & 0.1613 & 0.4711 & 0.2917  \\
      Llama-3.2-3B & 0.0772 & 0.2187 & \textbf{0.0898} & 0.3222 & 0.1770 & 0.2124 & 0.3493 & \textbf{0.1731} & 0.4581 & 0.2982  \\
      Llama-3.1-8B-Instruct & 0.0506 & 0.1869 & 0.0267 & 0.2989 & 0.1408 & 0.1496 & 0.2965 & 0.0602 & 0.4219 &  0.2321 \\
      \midrule
      Qwen-2.5-3B-Instruct  & 0.0631 & 0.1934 & 0.0374 & 0.3822 & 0.1690 & 0.1677 & 0.3056 & 0.084 & 0.5124 & 0.2674 \\
      Qwen-3-8B & 0.0524 & 0.2022 & 0.0372 & 0.3925 & 0.1711 & 0.1444 & 0.3117 & 0.0753 & 0.528 & 0.2649 \\
      \midrule
      FineEdit-X  & 0.0843 & 0.2077 & 0.0515 & 0.3379 & 0.1704 & 0.2077 & 0.3106 & 0.1006 & 0.4704 & 0.2723 \\
      FineEdit-Pro & \textbf{0.0914} & 0.2500 & 0.0511 & 0.4446 & 0.2093 & \textbf{0.2554} & 0.3729 & 0.1008 & 0.5858 & 0.3287 \\
      \midrule
      \rowcolor{yellow!20} HyperEdit (qwen-2.5-3B) & 0.0809 & \textbf{0.3894} & 0.0547 & \textbf{0.4665} & \textbf{0.2479} & 0.2132 & \textbf{0.5182} & 0.1084 & \textbf{0.6083} & \textbf{0.3620} \\
      \bottomrule
    \end{tabular}
  }
  \caption{Comparison of models on Diff-BLEU and Diff-ROUGE-L across domains in the Multi-turn Editing Task. Models are grouped by family and ordered by size.}
  \label{tab:diff-multiturn}
\end{table*}

Editing-specialized models show improved robustness. FineEdit-Pro achieves the best performance among existing baselines with 0.2093 Diff-BLEU and 0.3287 Diff-ROUGE-L, demonstrating that task-specific training helps models track revision dependencies across multiple editing rounds.

HyperEdit substantially outperforms all baselines in multi-turn settings. Our model achieves 0.2479 Diff-BLEU and 0.3620 Diff-ROUGE-L overall—an 18\% relative improvement over FineEdit-Pro. This larger gain compared to single-turn results (9\%) suggests that dynamic adaptation is particularly valuable when edits accumulate: the hypernetwork generates fresh, request-specific parameters for each round, preventing error propagation while maintaining awareness of both the current instruction and the evolving document state.

The advantages are especially pronounced on structured and long-form domains. On LaTeX, HyperEdit achieves 0.3894 / 0.5182, far exceeding FineEdit-Pro (0.2500 / 0.3729), indicating strong capability for handling complex structural revisions across multiple steps. On Wikipedia, HyperEdit reaches 0.4665 / 0.6083 compared to FineEdit-Pro's 0.4446 / 0.5858, demonstrating robust performance on lengthy, coherent text that requires maintaining consistency across edits. Even on Code and DSL, where precision is critical, HyperEdit maintains competitive or superior performance.

These results confirm that dynamic parameter generation enables more effective multi-turn editing by adapting to each instruction while preserving the integrity of previous modifications—a key requirement for practical iterative editing workflows. 
\begin{figure}[t]
    \centering
    \begin{subfigure}[b]{0.8\columnwidth}
        \centering
        \includegraphics[width=\textwidth]{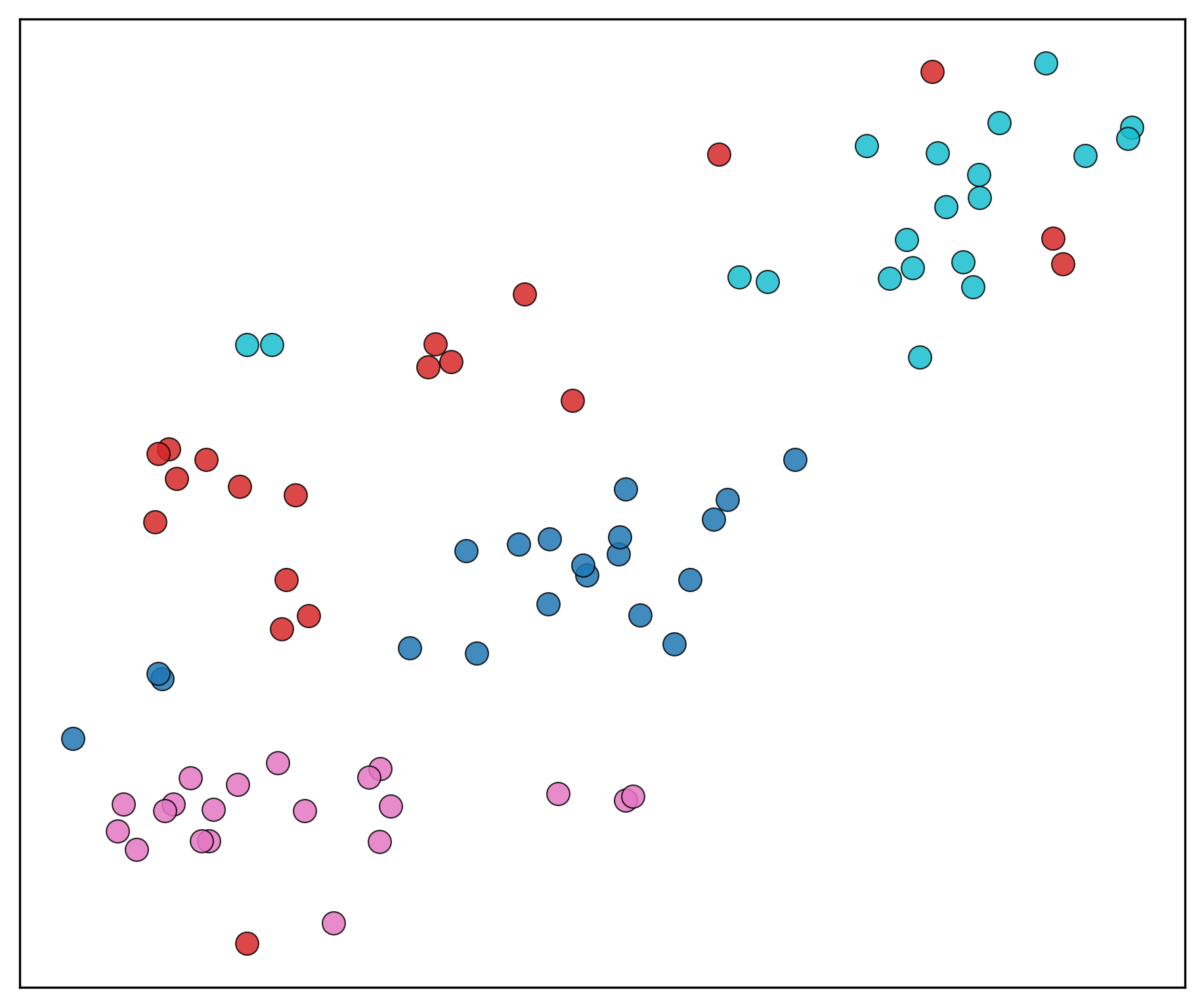}
        \caption{$\mathbf{A}^l$}
        \label{fig:tsne:a}
    \end{subfigure}

    \vspace{0.5em} 

    \begin{subfigure}[b]{0.8\columnwidth}
        \centering
        \includegraphics[width=\textwidth]{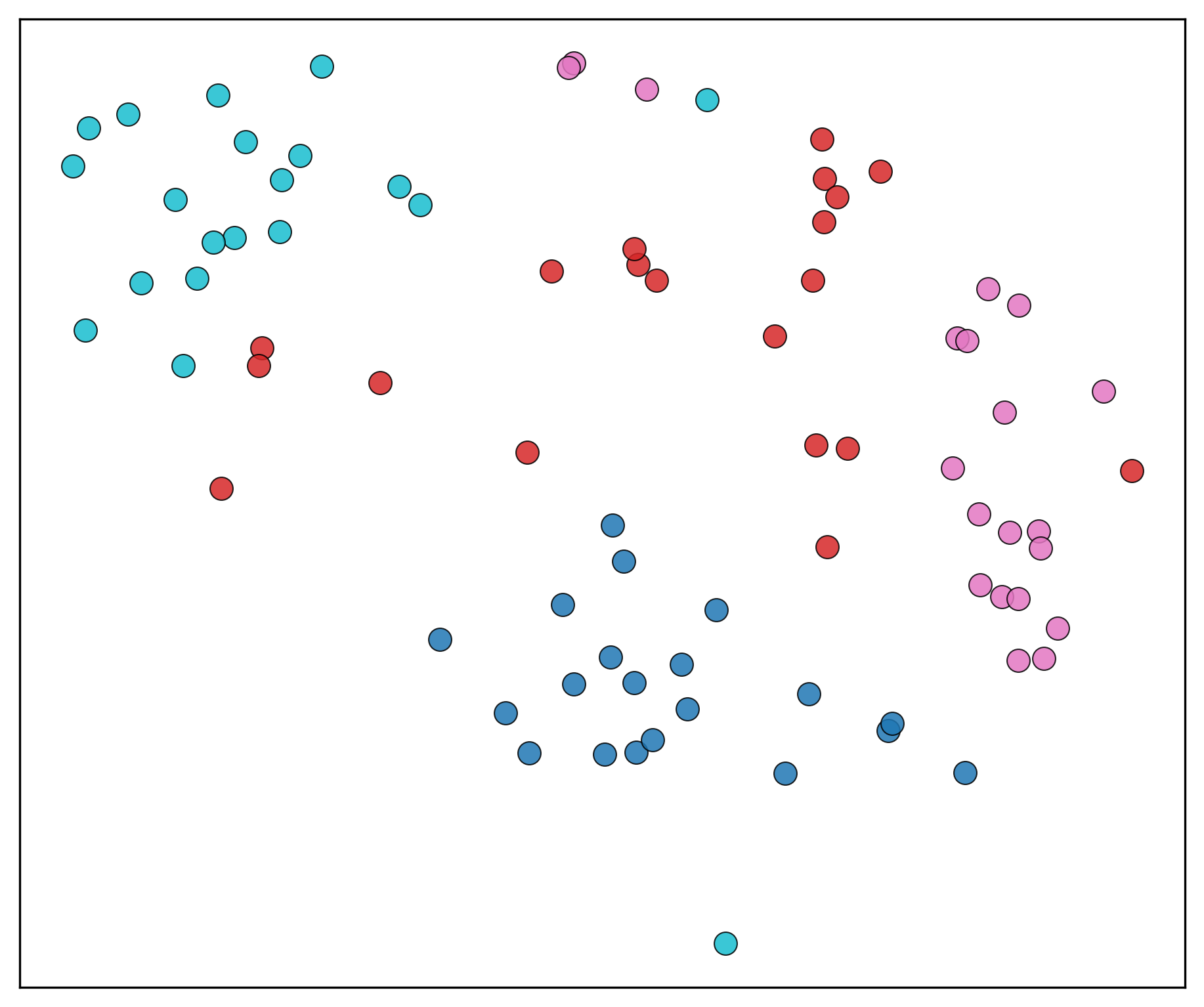}
        \caption{$\mathbf{A}^l\mathbf{B}^l$}
        \label{fig:tsne:b}
    \end{subfigure}

    \caption{t-SNE visualization of dynamic weight distributions across domains.
    Different editing domains are represented by the following colors: \textcolor{myblue}{\textbf{Code}},
\textcolor{myred}{\textbf{LaTeX}},
\textcolor{mypink}{\textbf{SQL}},
\textcolor{mycyan}{\textbf{Wiki}}.}
    \label{fig:tsne}
    \vspace{-5pt}
\end{figure}

\begin{figure}[t]
    \centering
    \includegraphics[width=0.9\columnwidth]{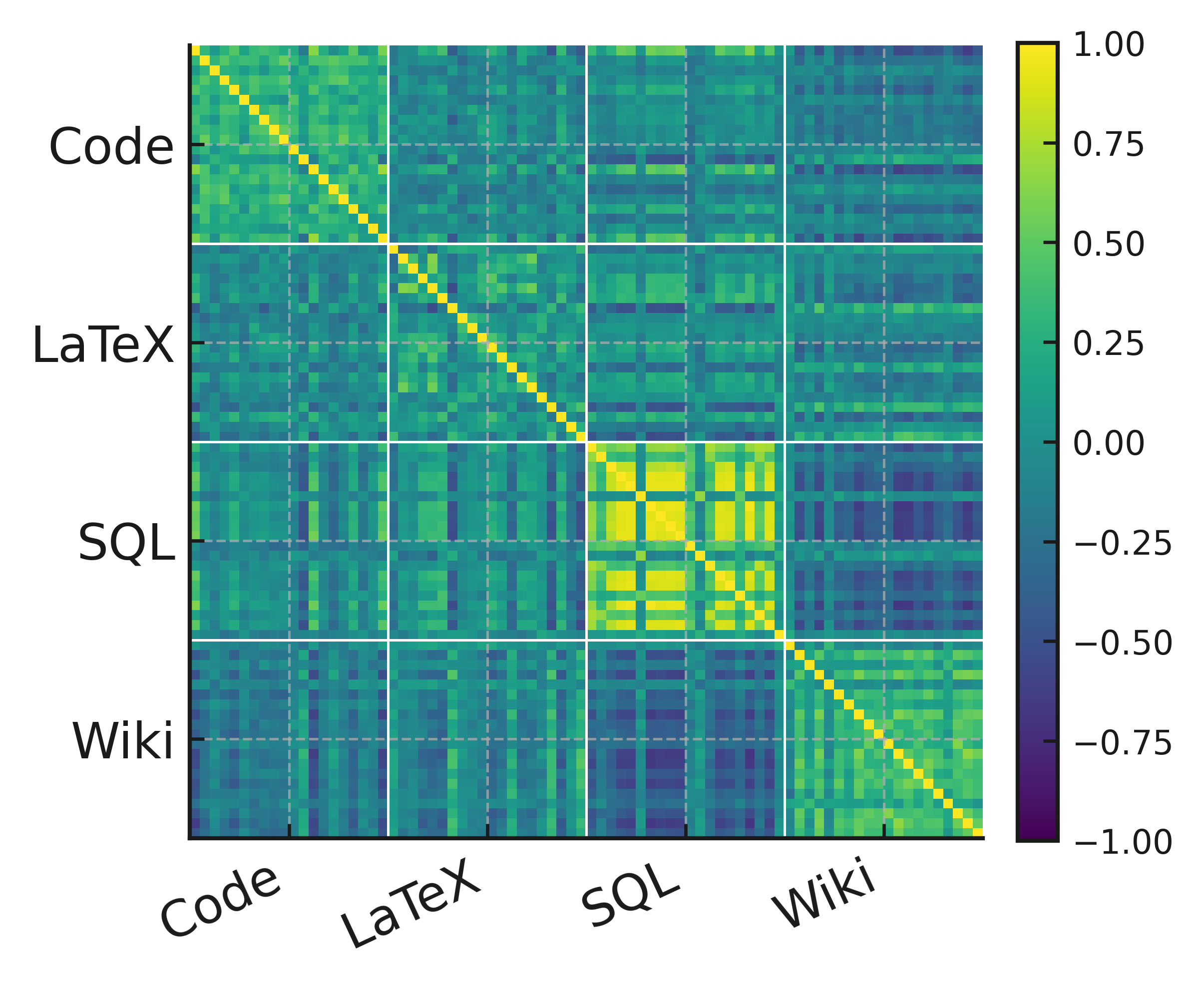}
    \caption{Dynamic Weight Similarity Matrix across editing domains. Pairwise cosine similarity of $\mathbf{A}^l$ Frobenius norms across 80 cases (20 per domain). Dark diagonal blocks show high intra-domain similarity; lighter off-diagonal regions indicate lower inter-domain similarity.}
    \label{fig:matrice}
    \vspace{-5pt}
\end{figure}

\subsection{Qualitative Study}
To verify that HyperEdit's hypernetwork generates meaningfully different parameters for different domains and editing scenarios, we analyze the generated dynamic weights. We randomly sampled 20 inference cases from each of four domains (\textit{Code}, \textit{LaTeX}, \textit{DSL}, and \textit{Wiki}), extracted the adapter matrices $\mathbf{A}^l$ and $\mathbf{B}^l$, computed their layer-wise Frobenius norms, and applied t-SNE for visualization.

Figure~\ref{fig:tsne:a} shows the $\mathbf{A}^l$ visualization, which reveals distinct, well-separated clusters for each domain. This confirms that $\mathbf{A}^l$, the hypernetwork's output, captures domain-specific editing strategies. \textit{Code} and \textit{DSL} form tight clusters, consistent with their reliance on rigid, syntax-driven transformations. \textit{LaTeX} and \textit{Wiki} exhibit more dispersed patterns, reflecting the greater linguistic and structural diversity in these domains.

In contrast, Figure~\ref{fig:tsne:b} shows the $\mathbf{A}^l\mathbf{B}^l$ representation with considerable overlap, especially between \textit{LaTeX} and \textit{Wiki}. This suggests that $\mathbf{B}^l$, the static low-rank projection, serves as a shared basis that enables cross-domain generalization while $\mathbf{A}^l$ provides domain-specific adaptation.

To quantify domain separation, we computed pairwise cosine similarities between the layer-wise Frobenius norms of $\mathbf{A}^l$ across all 80 cases. The resulting dynamic weight similarity matrix (Figure~\ref{fig:matrice}) exhibits clear block-diagonal structure with strong intra-domain similarity (dark blocks) and weaker inter-domain similarity (lighter off-diagonal regions). The average intra-domain similarity is 1.49× higher than inter-domain similarity, confirming that the hypernetwork consistently generates domain-specific weight patterns tailored to each editing context.

\subsection{Ablation Study}
To evaluate each component's contribution, we conduct an ablation study with three variants: (1) baseline with standard LoRA (no hypernetwork, no difference-aware loss), (2) hypernetwork only (dynamic adaptation without difference-aware loss), and (3) full HyperEdit (both components). All variants use identical  settings for fair comparison.

Table~\ref{tab:ablation} shows that the hypernetwork alone provides substantial improvements. Comparing variants (1) and (2), dynamic adaptation yields gains across most domains, particularly on structured tasks: LaTeX improves from 0.1278 to 0.1339 BLEU, and DSL from 0.0356 to 0.0521. On Wiki, the gains are even larger (0.3617 to 0.3423 BLEU), demonstrating that request-specific parameter generation helps the model adapt to diverse editing intents even without explicit supervision on modified regions.

Adding difference-aware regularization (variant 3) further improves performance. The full model achieves the best overall scores. The gains are most pronounced on LaTeX (0.1706 BLEU) and Wiki (0.3527 BLEU), where precise localization of edits is critical. On Code, the hypernetwork-only variant performs slightly better (0.0504 vs. 0.0492), likely because code editing involves more predictable, syntax-driven changes where aggressive localization may overly constrain the model. Overall, the full model achieves the best balance, confirming that both components contribute complementary benefits: dynamic adaptation for intent alignment and difference-aware regularization for local fidelity.

\section{Conclusion}
This paper proposes HyperEdit, a dual-modeling framework for instruction-based editing tasks. It adaptively aligns editing strategies with user intent through dynamic parameter generation driven by a hypernetwork. Combined with a difference-aware regularization, it ensures that modifications are focused on essential areas while preserving the original text as much as possible. Experimental results demonstrate that HyperEdit significantly outperforms general-purpose models with larger parameters and existing specialized editing methods, demonstrating its advantages in accuracy and local fidelity.

\clearpage

\section*{Limitations}
\noindent \textbf{Multi-turn Training.} Our current framework does not incorporate explicit multi-round training. Extending HyperEdit to support iterative optimization across multiple editing turns remains an important direction for future work.

\noindent \textbf{Computational Costs of Multi-Turn Evaluation.} Multi-turn evaluation requires repeated interactions, which significantly increase inference time and computational cost. Therefore, we did not include results from larger-scale models, such as 14B parameters or above, in this study.



\bibliography{anthology,custom}
\clearpage
\newpage

\appendix

\section{Parameter Settings}
\noindent \textbf{Chunking long context:}
Many large language models impose a fixed maximum token length \( L \) on their input (and sometimes output) sequences. Consequently, if the combination of \( T_{\text{orig}} \) and \( I_{\text{edit}} \) exceeds this limit, we divide the \( T_{\text{orig}} \) into smaller chunks of size \( \leq L \). Each chunk is then processed independently—paired with the same edit request and later concatenated to form the complete edited text. This approach ensures that every chunk fits within the model’s token budget, preventing overflow and reducing memory usage while preserving the overall structured editing behavior.

\noindent \textbf{Fine-Tuning Strategy:}  
We use Low-Rank Adaptation (LoRA) ~\cite{hu2022lora} to efficiently adapt these models to our task, significantly reducing the number of trainable parameters while preserving their expressive power. In all LoRA configurations, we set the rank $r=8$ and scaling $\alpha=32$, and use a dropout probability of 0.05. For both Llama-based and Qwen-based models, we apply  LoRA to the attention's projection layers through trainable low-rank matrices. We used the AdamW optimizer with a learning rate of $2 \times 10^{-5}$, training for 2 epochs, and set the effective batch size of 1 with gradient accumulation steps of 4 due to device limits. This strategy not only reduces computational overhead but also enables rapid convergence on our structured editing tasks. Preliminary experiments guided the choice of hyperparameters across all three model variants. 

\noindent \textbf{Decoding and Inference:}  
During generation, we set the temperature to 0.7 and used top-p sampling with a probability of 0.9 to balance diversity and coherence. Greedy decoding is applied by default if without sampling setting. The final edited text is obtained by merging the edited outputs from all chunks.
\section{Training Loss}

\begin{figure}[t]
    \centering
    \begin{subfigure}[b]{0.8\columnwidth}
        \centering
        \includegraphics[width=\textwidth]{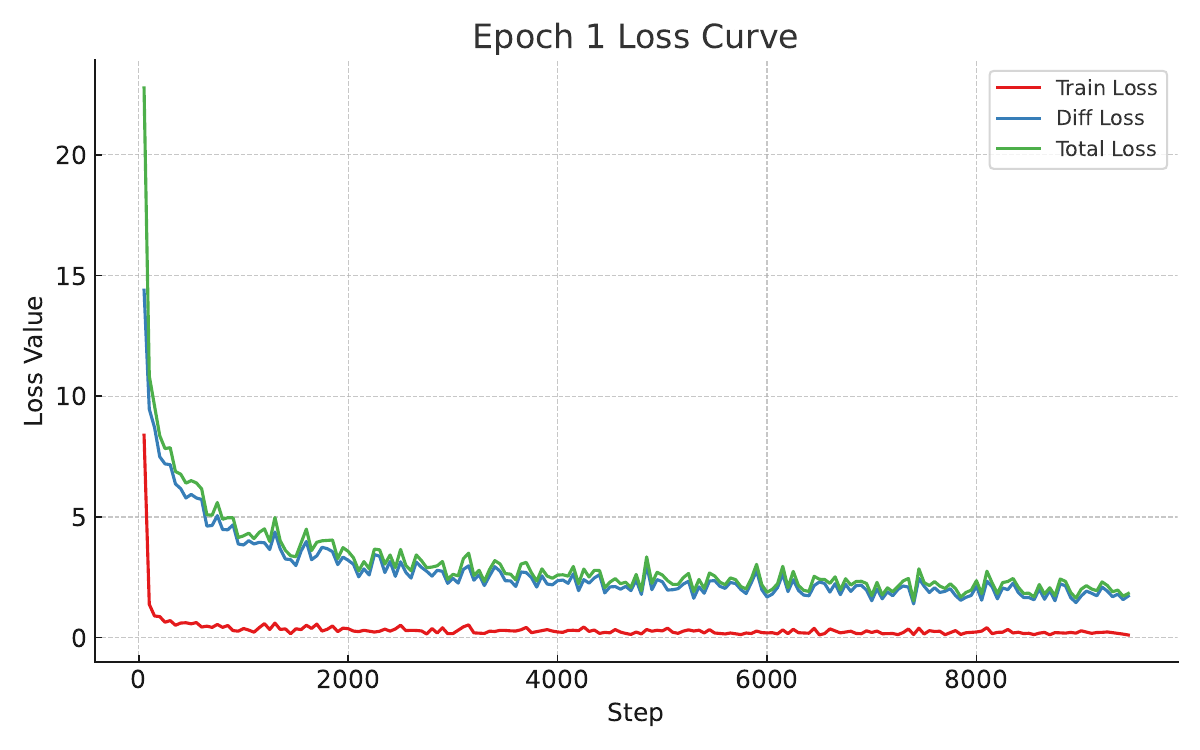}
        \caption{Training loss for Epoch 1}
    \end{subfigure}
    
    \vspace{0.5em} 
    
    \begin{subfigure}[b]{0.8\columnwidth}
        \centering
        \includegraphics[width=\textwidth]{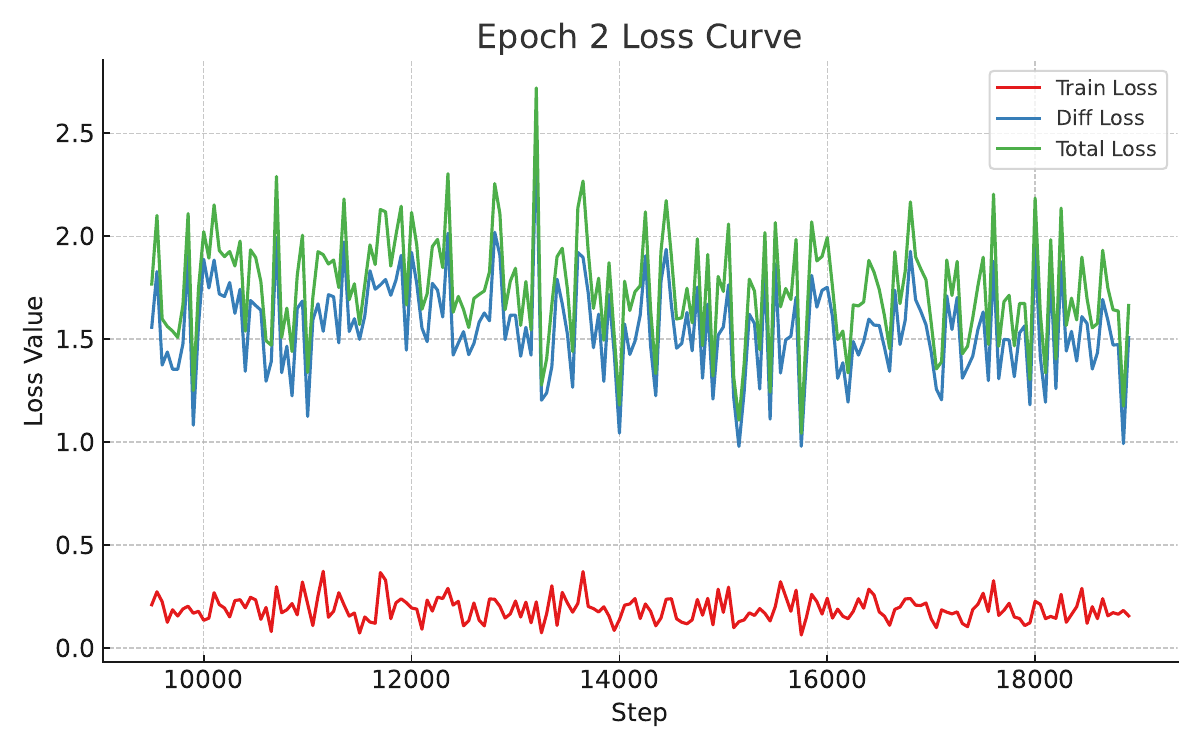}
        \caption{Training loss for Epoch 2}
    \end{subfigure}

    \begin{subfigure}[b]{0.8\columnwidth}
        \centering
        \includegraphics[width=\textwidth]{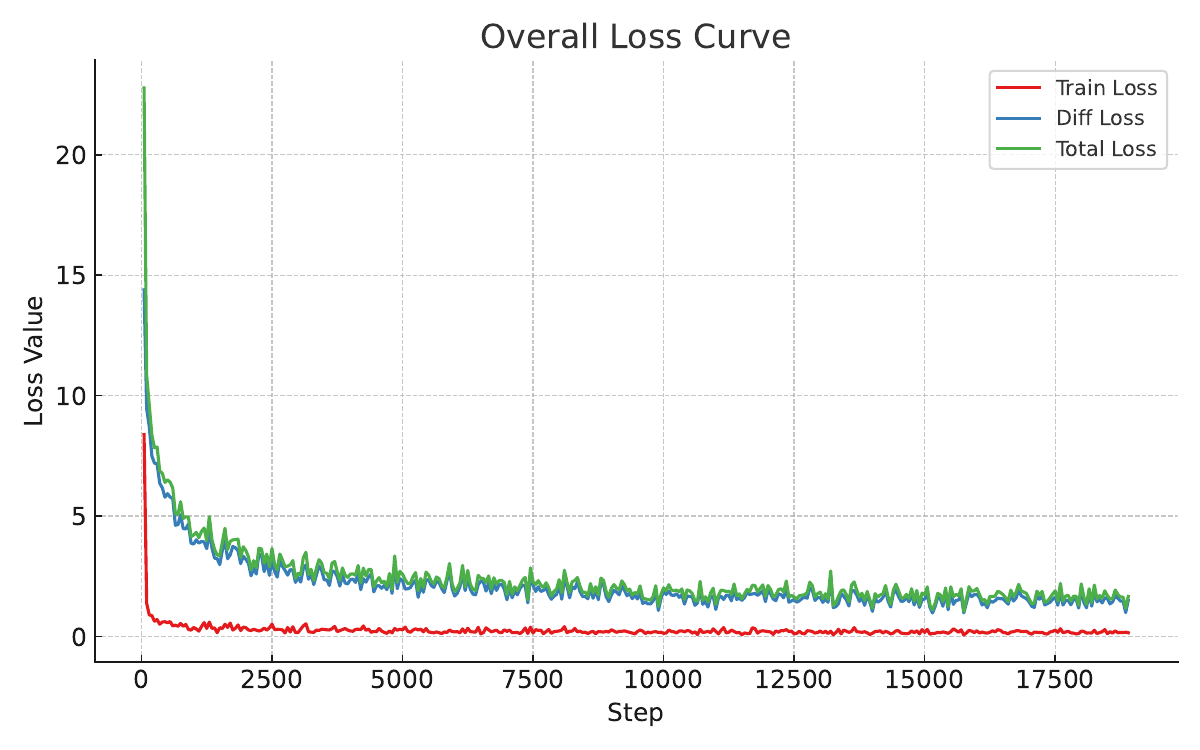}
        \caption{Overall Training Loss}
    \end{subfigure}
    \caption{Training loss for HyperEdit}

    \label{fig:loss}
    \vspace{-10pt}
\end{figure}

\noindent\textbf{Overview.} 
Figure~\ref{fig:loss} presents the evolution of the training losses during model optimization, 
including the training loss, diffusion loss, and total loss. 
The curves are displayed separately for Epoch~1, Epoch~2, and the overall training process.

\noindent\textbf{Epoch 1.} 
At the start of training, all loss components are relatively large, indicating that the model parameters 
are far from an optimal configuration. A rapid decline is observed within the first few hundred steps, 
reflecting efficient early learning of the data distribution. 
After approximately 2{,}000--3{,}000 steps, the losses begin to stabilize, 
suggesting that the model gradually approaches convergence. 
The total loss follows the same downward trend, showing consistent optimization across loss components.

\noindent\textbf{Epoch 2.} 
During the second epoch, the overall magnitude of the losses remains much lower compared with Epoch~1. 
The training loss becomes smoother, showing only small fluctuations. 
This indicates that the model mainly performs fine adjustments to the learned representations 
rather than major updates. 
The diffusion loss also remains stable within a narrow range, suggesting that the optimization process 
has reached a steady state.


\section{Efficiency and Computational Cost Analysis}
\label{app:efficiency}

To comprehensively evaluate the efficiency of HyperEdit compared to baselines, we conducted a detailed analysis of runtime latency, memory consumption, and theoretical computational complexity (FLOPs).

We measured the inference latency and peak memory usage on a single NVIDIA RTX A6000 GPU. For each domain (Code, LaTeX, DSL, Wikipedia), we randomly sampled 100 instances. The results, summarized in Table \ref{tab:runtime}, compare the base model (Qwen-2.5-Instruct-3B), FineEdit-Pro, and HyperEdit.

\paragraph{Runtime Analysis.}
As shown in Table \ref{tab:runtime}, the base model averages 32.17 seconds per instance. FineEdit-Pro, which utilizes standard LoRA, averages 53.15 seconds. HyperEdit averages 53.83 seconds, showing a negligible increase over FineEdit-Pro. Despite introducing a hypernetwork to generate request-specific parameters, HyperEdit's inference latency remains comparable to the static LoRA baseline. This is because the parameter generation occurs once per request (or is parallelizable), and the decoding process dominates the total runtime.

\paragraph{Memory Footprint.}
In terms of memory usage, the base model exhibits a peak of 9.96GB, while FineEdit-Pro reaches 10.6GB. HyperEdit requires a peak memory of 18.7GB. This increase is expected, as HyperEdit maintains dynamic, instance-specific adapters for each layer in memory during inference, unlike standard LoRA which can share or merge static weights. 

\begin{table}[h]
\centering
\resizebox{\linewidth}{!}{
\begin{tabular}{l|ccc|c}
\hline
\textbf{Category} & \textbf{Qwen-2.5-instruct-3B} & \textbf{FineEdit-Pro} & \textbf{HyperEdit} & \textbf{Count} \\
\hline
Code & 20.61s & 33.33s & 37.49s & 100 \\
LaTeX & 30.70s & 48.34s & 56.22s & 100 \\
DSL & 47.17s & 78.27s & 64.42s & 100 \\
Wikipedia & 30.21s & 52.66s & 57.20s & 100 \\
\hline
\textbf{Average} & \textbf{32.17s} & \textbf{53.15s} & \textbf{53.83s} & - \\
\textbf{Peak Memory} & \textbf{9.96GB} & \textbf{10.60GB} & \textbf{18.70GB} & - \\
\hline
\end{tabular}
} 
\caption{Average inference time per instance across different domains. All models are evaluated on the same NVIDIA A6000 GPU.}
\label{tab:runtime}
\end{table}
\paragraph{Theoretical FLOPs Analysis}
We further analyze the theoretical floating-point operations (FLOPs) to validate the efficiency of our approach.

The backbone for all models is Qwen-2.5-Instruct-3B, which contains approximately 3 billion parameters. For decoder-only Transformers, the inference cost is widely estimated as $2 \times N$ FLOPs per token, where $N$ is the parameter count \cite{casson2023transformerflops}. Thus, the baseline cost is approximately $2 \times 3 \times 10^9 = 6$ GFLOPs per token. Standard LoRA introduces low-rank matrices with a computational overhead of $O(r/d)$, where $r$ is the rank and $d$ is the hidden dimension. Since $r \ll d$, this overhead is typically less than 1\%, keeping the inference cost effectively unchanged at $\approx 6$ GFLOPs per token.

HyperEdit builds upon the 3B-parameter backbone. Its additional components include a SentenceTransformer encoder and a GRU-based HyperLoRA module.

\paragraph{Hypernetwork Overhead:} The hypernetwork generates the adapter weights. This process involves the sentence encoder (executed once per prompt) and the weight generation network. The complexity of generating rank-$r$ updates is proportional to $O(r/d)$, which is computationally lightweight compared to the dense projections of the backbone.
paragraph{Inference Cost:} Once weights are generated, the token-by-token decoding follows the same path as standard LoRA.

Consequently, the total inference complexity of HyperEdit is dominated by the backbone model. The per-token cost remains approximately 6 GFLOPs, with only marginal overhead incurred during the encoding phase. This confirms that HyperEdit achieves dynamic adaptation without a significant penalty in computational throughput.

\begin{algorithm}[h]
\caption{Hierarchical Edited Span Extraction Algorithm}
\label{alg:span_extraction}
\SetKwInput{Input}{Input}
\SetKwInput{Output}{Output}
\SetKwFunction{Normalize}{Normalize}
\SetKwFunction{Align}{SequenceMatcher}
\SetKwFunction{FineGrained}{FineGrainedRefinement}
\SetKwFunction{Mark}{MarkDelete}

\Input{Reference text $Y$, Candidate text $\hat{Y}$}
\Output{Reference Span Set $Y_{\text{edit}}^{\text{span}}$, Candidate Span Set $\hat{Y}_{\text{edit}}^{\text{span}}$}

\tcp{Step 1: Semantic Normalization}
$\tilde{Y} \leftarrow \Normalize(Y)$\;
$\tilde{\hat{Y}} \leftarrow \Normalize(\hat{Y})$\;

\tcp{Step 2: Coarse-Grained Line Alignment}
$\mathcal{M} \leftarrow \Align(\tilde{Y}, \tilde{\hat{Y}})$ \tcp*{Get opcodes: equal, insert, delete, replace}
$Y_{\text{edit}}^{\text{span}} \leftarrow \epsilon, \quad \hat{Y}_{\text{edit}}^{\text{span}} \leftarrow \epsilon$\;

\tcp{Step 3: Iterative Processing \& Fine-Grained Refinement}  
\ForEach{operation tuple $m = \langle tag, i_1, i_2, j_1, j_2 \rangle \in \mathcal{M}$}{
    \uIf{$tag = \texttt{equal}$}{
        \textbf{continue} \tcp*{Discard invariant context}
    }
    \uElseIf{$tag = \texttt{insert}$}{
        \tcp{Append inserted segments}
        $Y_{\text{edit}}^{\text{span}} \leftarrow Y_{\text{edit}}^{\text{span}} \oplus \tilde{\hat{Y}}[j_1:j_2]$\;
        $\hat{Y}_{\text{edit}}^{\text{span}} \leftarrow \hat{Y}_{\text{edit}}^{\text{span}} \oplus \tilde{\hat{Y}}[j_1:j_2]$\;
    }
    \uElseIf{$tag = \texttt{delete}$}{
        \tcp{Append marked deletions}
        $del\_seg \leftarrow \Mark(\tilde{Y}[i_1:i_2])$\;
        $Y_{\text{edit}}^{\text{span}} \leftarrow Y_{\text{edit}}^{\text{span}} \oplus del\_seg$\;
        $\hat{Y}_{\text{edit}}^{\text{span}} \leftarrow \hat{Y}_{\text{edit}}^{\text{span}} \oplus del\_seg$\;
    }
    \uElseIf{$tag = \texttt{replace}$}{
        \tcp{Step 4: Recursive Character-Level Refinement}
        $(t_{old}, t_{new}) \leftarrow \FineGrained(\tilde{Y}[i_1:i_2], \tilde{\hat{Y}}[j_1:j_2])$\;
        $Y_{\text{edit}}^{\text{span}} \leftarrow Y_{\text{edit}}^{\text{span}} \oplus t_{new}$ \tcp*{Target content}
        $\hat{Y}_{\text{edit}}^{\text{span}} \leftarrow \hat{Y}_{\text{edit}}^{\text{span}} \oplus t_{new}$ \tcp*{Generated content}
    }
}

\Return{$Y_{\text{edit}}^{\text{span}}, \hat{Y}_{\text{edit}}^{\text{span}}$}
\end{algorithm}

\section{Implementation of Edited Span Extraction}
\label{app:span_extraction}

The validity of our proposed metrics, Diff-BLEU and Diff-ROUGE-L, relies on the accurate isolation of edited spans from the invariant context. To achieve this, we developed a deterministic, hierarchical extraction algorithm ~\ref{alg:span_extraction} that processes reference and candidate texts through a multi-stage pipeline. This process ensures that the evaluation focuses exclusively on semantic modifications while robustly handling formatting noise. 

\paragraph{Semantic Normalization.}
Raw textual comparisons are often sensitive to trivial formatting variations, such as indentation changes or trailing whitespace, which do not reflect semantic errors in code or natural language. To mitigate this, we apply a regular expression-based normalization to both the reference ground truth ($Y$) and the model prediction ($\hat{Y}$). This operation collapses consecutive whitespace characters into a single space and strips boundary voids, ensuring that the subsequent alignment is driven by token identity rather than stylistic artifacts.

\paragraph{Coarse-Grained Line Alignment.}
We utilize the \texttt{SequenceMatcher} algorithm (based on the Ratcliff-Obershelp pattern matching) to align the normalized texts at the line level. The algorithm identifies the Longest Common Subsequence (LCS) of lines and categorizes textual segments into four distinct operation codes:
\begin{itemize}
    \item \texttt{equal}: Contextual lines present in both sequences. These are immediately discarded to prevent the inflation of metrics by unchanged content.
    \item \texttt{delete}: Lines present in $Y$ but absent in $\hat{Y}$.
    \item \texttt{insert}: Lines present in $\hat{Y}$ but absent in $Y$.
    \item \texttt{replace}: Blocks of lines where the content has been modified but corresponds positionally.
\end{itemize}

\paragraph{Fine-Grained Span Refinement.}
For segments tagged as \textit{replace}, a simplistic line-level extraction would be imprecise if only a single token within a long line was altered. 
To address this, our algorithm performs a recursive, character-level alignment strictly within the replace blocks.
\begin{itemize}
    \item We isolate the specific sub-sequences within the line that differ, distinguishing between original tokens ($t_{old}$) and target tokens ($t_{new}$).
    \item This ensures that even within a modified line, unchanged distinct tokens are excluded from the evaluation spans, maximizing the sensitivity of the metric.
\end{itemize}

\paragraph{Span Aggregation for Metric Calculation.}
Finally, we aggregate the extracted components to construct the evaluation sets. 
The reference span set $Y_{\text{edit}}^{\text{span}}$ is constructed by concatenating: 
(1) segments marked as \textit{insert}, 
(2) the $t_{new}$ components from \textit{replace} operations, and 
(3) the \textit{delete} segments (represented explicitly with special markers, e.g., \texttt{[-token-]}).
Similarly, the candidate span set $\hat{Y}_{\text{edit}}^{\text{span}}$ is formed by concatenating the corresponding model-predicted segments. 
By including explicit representations of deletions, our metric evaluates not only the quality of generated text but also the precision of deletion operations. 
These aggregated spans serve as the direct inputs for Eq.~(10) and Eq.~(11).

\end{document}